\documentclass[twoside,11pt]{article}

\usepackage{blindtext}
\usepackage[abbrvbib, preprint]{jmlr2e}
\usepackage{array}
\usepackage{booktabs}
\usepackage{makecell}
\usepackage[misc]{ifsym}
\usepackage{booktabs}
\usepackage{multirow}
\usepackage{hyperref} 

\hypersetup{
    colorlinks=true,      
    citecolor=blue,       
    linkcolor=blue,       
    urlcolor=blue         
}

\usepackage{lastpage}
\jmlrheading{23}{2026}{1-\pageref{LastPage}}{1/21; Revised 5/22}{9/22}{21-0000}{Correspondence to \textit{leiyo@dtu.dk}}

\firstpageno{1}

\begin{document}

\title{\texttt{xai-cola}: A python library for sparsifying counterfactual explanations}


\author{\name Lin Zhu \email s232291@student.dtu.dk\\
\name Lei You \textnormal{\textsuperscript\Letter} \email leiyo@dtu.dk\\
\addr Technical University of Denmark, Copenhagen, Denmark}

\editor{My editor}

\maketitle

\begin{abstract}
Counterfactual explanation (CE) is an important domain within post-hoc explainability. However, the explanations generated by most CE generators are often highly redundant. This work introduces an open-source Python library \texttt{xai-cola}, which provides an end-to-end pipeline for sparsifying CEs produced by arbitrary generators, reducing superfluous feature changes while preserving their validity. It offers a documented API that takes as input raw tabular data in \texttt{pandas} \texttt{DataFrame} form, a preprocessing object (for standardization and encoding), and a trained \texttt{scikit-learn} or \texttt{PyTorch} model. On this basis, users can either employ the built-in or externally imported CE generators. The library also implements several sparsification policies and includes visualization routines for analysing and comparing sparsified counterfactuals. \texttt{xai-cola} is released under the MIT license and can be installed from PyPI. Empirical experiments indicate that \texttt{xai-cola} produces  sparser counterfactuals across several CE generators, reducing the number of modified features by up to 50\% in our setting. The source code is available at \url{https://github.com/understanding-ml/COLA}.

\end{abstract}


\section{Introduction}
Explainable artificial intelligence (XAI) has become a critical topic as complex black-box models are increasingly deployed in socially sensitive and safety-critical domains. XAI methods aim to make model behavior more transparent and understandable, supporting trust, debugging, and regulatory compliance. Broadly, XAI research is divided into \emph{Ante-hoc} (transparent models) and \emph{Post-hoc} explainability. The former is delineated into simulatability, decomposability, and algorithmic transparency, whereas the latter is categorized into model-agnostic and model-specific approaches, encompassing sub-domains such as feature attribution, permutation feature importance, and local explanations~\citep{arrieta2020explainable}.

Within \emph{Post-hoc} explainability, counterfactual explanation (CE) has emerged as a prominent approach, which shows how small changes in input features can flip a model’s prediction~\citep{wachter2017counterfactual}. Due to diverse constraints and goals, hundreds of algorithms for generating CEs have been proposed~\citep{guidotti2024counterfactual,verma2024counterfactual}. Since our focus is on how these algorithms produce counterfactual examples given a model and data, we refer to them as CE generators in the remainder of this paper. Mainstream CE generators can be roughly divided into: (i) instance-level CE generators that aim for diverse and plausible counterfactual explanations~\citep{poyiadzi2020face, mothilal2020explaining}; and (ii) global- or distribution-level CE generators that indicate how to move a group of factual instances to achieve desired model outputs~\citep{rawal2020beyond,carrizosa2024mathematical,pmlr-v258-you25a}. In practice, however, many CEs modify a large number of features, which can make them difficult to act on or even infeasible given domain constraints.

In this paper, we introduce \texttt{xai-cola}, an open-source Python library to sparsify CEs. Sparsification, in this context, refers to the process of reducing superfluous feature changes while preserving the validity of the counterfactuals. The goal of \texttt{xai-cola} is to provide a general pipeline that can be applied on top of existing CE generators to obtain sparser CEs. The core sparsification algorithm was proposed, called COLA~\citep{you2024refining}; this library implements that algorithm in a modular way and extends it with configurable components. It offers a unified interface, including a \texttt{Model} class for wrapping prediction models and a \texttt{COLAData} class for handling datasets, to ensure compatibility with a broad range of CE algorithms. The library also provides tailored visualization utilities for inspecting and comparing sparsified counterfactuals. In addition, to support an end-to-end workflow, \texttt{xai-cola} includes two built-in CE generators: an instance-level CE generator based on DiCE~\citep{mothilal2020explaining} and a distribution-level CE generator adapted from DisCount~\citep{pmlr-v258-you25a}, so that users can generate and sparsify CEs within a single pipeline.

\section{The \texttt{xai-cola} Library}
\label{sec:library}

\texttt{xai-cola} is implemented as a set of small, composable modules and is available on PyPI.\footnote{Documentation and installation instructions are available at \url{https://cola-docs.readthedocs.io/en/latest/}.}
The library is designed to be model-agnostic and easy to integrate with existing pipelines: users only need to provide tabular data, a preprocessing object, and a trained predictive model.
On top of these inputs, \texttt{xai-cola} offers four core components:
(i) a data container that stores factual and counterfactual instances together with metadata;
(ii) a model interface that wraps different learning backends under a unified API;
(iii) 2 wrapped built-in CE generators ; and
(iv) a sparsification and visualization layer that sparsify and inspect CEs.

\begin{figure}[ht]
    \centering
    \includegraphics[width=\linewidth]{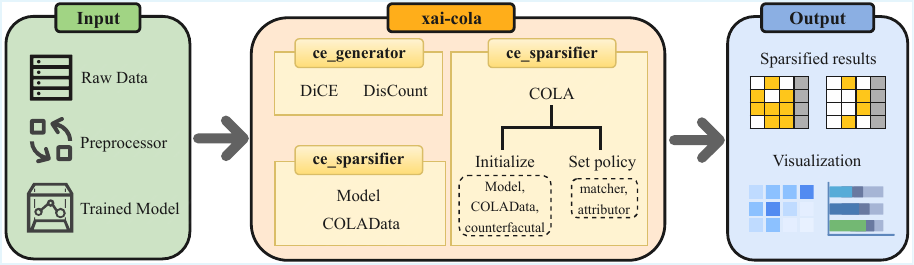}
    \caption{Overall architecture of the \texttt{xai-cola} pipeline.}
    \label{fig:framework}
\end{figure}

Figure~\ref{fig:framework} summarizes the end-to-end workflow.
On the left, the user provides raw tabular data, a preprocessing object (e.g., a \texttt{scikit-learn} pipeline that performs standardization and encoding), and a trained predictive model defined in the preprocessed feature space.
These inputs are wrapped by the data and model components in the central block.
The upper part of the central block corresponds to the \texttt{ce\_generator} subpackage, which can be used to generate CEs. The lower part corresponds to the \texttt{ce\_sparsifier} subpackage, which hosts the core classes for data and model handling as well as the main sparsification routine.
The right block shows the outputs: sparsified counterfactuals and visualization routines that summarize sparsity and validity properties.

In the package, these components are organized into two subpackages:
\texttt{ce\_generator}, which hosts built-in CE generators such as \texttt{DiCE} and \texttt{DisCount},
and \texttt{ce\_sparsifier}, which contains the core classes \texttt{Model}, \texttt{COLAData}, and \texttt{COLA} together with modules for data handling, sparsification policies, utilities, and visualization.

\paragraph{Model interface.}
The \texttt{Model} class wraps compatible \texttt{scikit-learn} and \texttt{PyTorch} models behind a unified interface.
Given a backend specification, it automatically detects the type of the underlying model and exposes a common set of operations:
prediction and class-probability queries, and gradient-based calls when available. This unified interface allows the \texttt{COLA} class and the built-in CE generators to invoke arbitrary predictive models in a consistent way.

\paragraph{Data container.}
The \texttt{COLAData} class provides a structured container for both factual and counterfactual data. It accepts inputs either as a \texttt{pandas} \texttt{DataFrame} or as a \texttt{NumPy} array together with column names, and records feature names, feature types (e.g., numerical vs.\ categorical), and the label column. These design choices make it possible to centralize all data-related operations inside \texttt{COLA} via a single, well-structured container.

\paragraph{CE generators.}
For generating CEs, \texttt{xai-cola} supports both external and built-in CE generators.
External generators are entirely implemented outside the library: the only requirement is that they return a \texttt{DataFrame} which can subsequently be provided as input to \texttt{COLAData}, which demonstrates the extensibility of our framework and the ease with existing CE generators.
For users who prefer to stay within the library, the \texttt{ce\_generator} subpackage provides two built-in generators, \texttt{DiCE} and \texttt{DisCount}.
The \texttt{DiCE} wrapper adapts the widely used instance-level DiCE generator~\citep{mothilal2020explaining} to the \texttt{Model} and \texttt{COLAData} interfaces, delegating the actual generation to the original implementation.
For \texttt{DisCount}, we refactor the original source code from~\citet{pmlr-v258-you25a} into a user-facing class with an interface compatible with \texttt{Model} and \texttt{COLAData}.
In both cases, the generated CEs will be inserted back into the \texttt{COLAData} object and then passed for sparsification and visualization.

\paragraph{Sparsification and visualization.}
The \texttt{COLA} class orchestrates the sparsification stage.
It takes as input a \texttt{Model} instance and a \texttt{COLAData} instance that already contains candidate counterfactuals.
Sparsification policies is configured via the method \texttt{set\_policy} where the library provides a finite set of predefined matchers methods and attributors that can be chosen by name. Once a policy has been chosen, sparsification is invoked by calling \texttt{get\_refined\_counterfactuals}, with an argument that controls the maximum number of features allowed to change in all CEs. This method performs a sequence of operations—including matching factual and counterfactual instances, computing feature attributions, and composing refined counterfactuals according to the chosen policy, and returns a \texttt{DataFrame} of sparsified counterfactuals. Finally, \texttt{COLA} offers convenience methods that interface with the visualization utilities in \texttt{ce\_sparsifier.visualization}, allowing users to inspect sparsity patterns and compare different policies through standardized plotting routines.

\vspace{-2mm}
\section{Experiments}

To the best of our knowledge, this is the first Python library to sparsify CEs. As a result, there are no directly comparable prior benchmarks, and our experiments focus on assessing the degree to which \texttt{xai-cola} sparsifies counterfactuals generated by different CE generators. We evaluate our approach on a diverse set of established counterfactual generators: Wachter \citep{wachter2017counterfactual}, DiCE \citep{mothilal2020explaining}, GLOBE-CE \citep{ley2023globe}, and DisCount \citep{pmlr-v258-you25a}.
\vspace{-2mm}
\begin{table}[ht]
    \centering
    \small
    \setlength{\tabcolsep}{4pt} 
    \renewcommand{\arraystretch}{0.95} 
    \caption{Sparsification performance of \texttt{xai-cola} across different CE generators on the German Credit and COMPAS datasets.}
    \vspace{-2mm}
    \resizebox{\linewidth}{!}{
    \begin{tabular}{@{}l c c cc cc@{}} 
        \toprule
        \multirow{2}{*}{CE generator} &
        \multirow{2}{*}{Type} &
        \multirow{2}{*}{Publication} &
        \multicolumn{2}{c}{German credit} &
        \multicolumn{2}{c}{COMPAS} \\
        \cmidrule(lr){4-5} \cmidrule(lr){6-7}
        & & & sklearn & pytorch & sklearn & pytorch \\
        \midrule
        Wachter   & instance-level      & Harv.\ JL \& Tech., 2017 & ---    & 10.4\%      & ---    & 13.8\%      \\
        DiCE      & instance-level      & FAccT, 2020              & 24.9\% & 12.0\%      & 11.2\% & 8.2\%       \\
        GLOBE-CE  & group-level         & ICML, 2023               & 9.2\%  & 8.4\%       & 14.4\% & 10.0\%      \\
        DisCount  & distribution-level  & AISTATS, 2025            & ---    & 30.3\%      & ---    & 45.3\%      \\
        \bottomrule
    \end{tabular}
    }
    \label{tab:ce_generators}
    \vspace{1mm}
    \footnotesize\emph{Note.} Wachter and DisCount require gradient-based optimization, not applicable to sklearn.
\end{table}

The experimental results are summarized in Table~\ref{tab:ce_generators}. As illustrated, \texttt{xai-cola} consistently yields sparser counterfactuals across all evaluated generators~\footnote{Experiments in this paper differ from the original algorithm. We compute (and report) Shapley attributions at the original feature level rather than on one-hot-expanded dimensions, ensuring that feature importance corresponds to user-facing variables and actionable edits, and avoiding attribution fragmentation caused by encoding.}. In particular, it achieves substantial reductions in the number of feature changes, decreasing the required modifications by up to 50\% in the best cases. For the predictive models used in Table~\ref{tab:ce_generators}, entries labeled “sklearn” correspond to logistic regression models implemented with scikit-learn~\citep{scikit-learn}, while “pytorch” refers to feed-forward neural networks implemented in PyTorch. We use standard configurations for these models without extensive hyperparameter tuning, since our primary goal is to demonstrate the relative sparsity improvements provided by our library rather than to maximize base model accuracy.
\vspace{-2mm}
\section{Conclusion}
This paper introduces \texttt{xai-cola}, an open-source Python library for sparsifying counterfactual explanations. The library provides a unified interface to predictive models and CE generators, an easy-to-use pipeline for integrating sparsification into existing workflows, and a diverse set of algorithmic and visualization options. In future work, we plan to extend the library with additional CE generators and sparsification strategies. Our goal is to make \texttt{xai-cola} a one-stop toolkit for generating, sparsifying, and analyzing CEs and we welcome contributors to help us build a better pipeline.

\newpage
\bibliography{sample}

\end{document}